\pdfoutput=1
\documentclass[journal,twoside,web]{packages/ieeecolor}
\usepackage{packages/jsen}

\usepackage{times}
\usepackage{epsfig}
\usepackage{graphicx}
\usepackage{amsmath}
\usepackage{amssymb}
\usepackage{amsfonts}
\usepackage{subfigure} 
\usepackage[ruled]{algorithm}
\usepackage{algpseudocode}
\usepackage{tikz}
\usepackage{ctable}
\usepackage[export]{adjustbox}
\usepackage{import}
\usepackage{pgf}
\usepackage[bookmarks=true]{hyperref}

\usepackage{ulem}
\usepackage{tcolorbox}
\usepackage{wrapfig}
\definecolor{abstractbg}{rgb}{0.89804,0.94510,0.83137}

\makeatletter
\let\NAT@parse\undefined
\makeatother
\usepackage[numbers]{natbib}

\usepackage{packages/sparo_acronyms}
\usepackage{packages/sparo_math}
\usepackage{packages/sparo_SIunits}
\usepackage{packages/sparo_misc}
\usepackage{multirow}

\usepackage{soul,color}
\usepackage{verbatim} %

\usepackage{xcolor}

\usepackage[symbol]{footmisc}

\title{Joint-ID: Transformer-based Joint Image Enhancement and Depth Estimation for Underwater Environments}

\author{Geonmo Yang${}^{1}$, Gilhwan Kang${}^{1}$, Juhui Lee${}^{1}$, Younggun Cho${}^{1*}$
\thanks{${}^{1}$ G. Yang, J. Lee, G, Kang and Y. Cho are with the Dept. Electr. and Comput. Eng., Inha University, S. Korea \texttt{(e-mail: [ygm7422, wngml6635, rlfghks527]@inha.edu, yg.cho@inha.ac.kr)} }
}%

\begin{document}
    
\IEEEtitleabstractindextext{%
\fcolorbox{abstractbg}{abstractbg}{%
\begin{minipage}
    {\textwidth}%
    \begin{wrapfigure}[16]{r}{3.8in}%
        \includegraphics[width=3.8in]{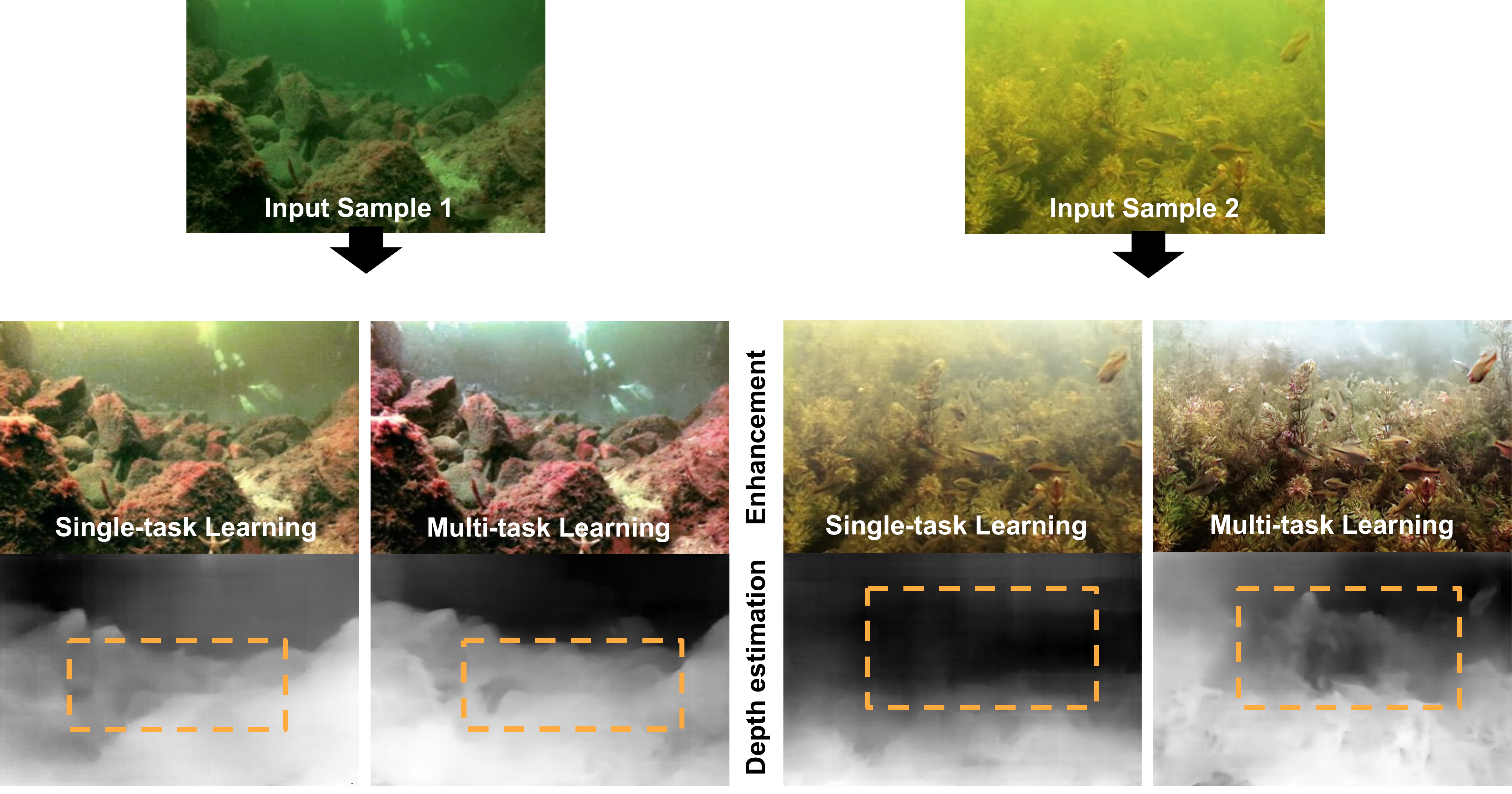}%
    \end{wrapfigure}%
\begin{abstract}

Underwater imaging is a challenging task due to factors such as scattering, absorption, and turbulence, which degrade image quality and limit visibility. In this paper, we propose a novel approach for enhancing underwater images that leverages the benefits of joint learning for simultaneous image enhancement and depth estimation. We introduce \texttt{Joint-ID}, a transformer-based neural network that can obtain high-perceptual image quality and depth information from raw underwater images. Our approach formulates a multi-modal objective function that addresses invalid depth, lack of sharpness, and image degradation based on color and local texture. We design an end-to-end training pipeline that enables joint restoration and depth estimation in a shared hierarchical feature space. Additionally, we propose a synthetic dataset with various distortions and scene depths for multi-task learning. We evaluate \texttt{Joint-ID} on synthetic and standard datasets, as well as real underwater images with diverse spectra and harsh turbidity, demonstrating its effectiveness for underwater image enhancement and depth estimation. Furthermore, we demonstrate the ability of Joint-ID to perform feature matching and saliency detection for visually guided underwater robots. Our proposed method has the potential to improve the visual perception of underwater environments and benefit applications such as oceanography and underwater robotics. Supplements are available at: \textit{\href{https://sites.google.com/view/joint-id/home}{https://sites.google.com/view/joint-id/home}}.

\end{abstract}

\end{minipage}}
}

\maketitle

\section{Introduction}

Image enhancement is essential for underwater investigations and applications that rely on autonomous robot vision, such as ecological surveys, robot salvage, crack management, and
resource exploration. Underwater vision suffers from more complex distortion problems than ground vision, including poor visibility, color distortion, light scattering, absorption, and hazing, which hinder the navigation of autonomous underwater robots.
To address these challenges, researchers have actively investigated various image enhancement techniques to enable more accurate perception and navigation. 

Underwater light propagation can be modeled using physical principles, which opens up opportunities for physics-based image enhancement methods \cite{berman2017diving, cho2016online, fu2014retinex, galdran2015automatic}. Underwater images often exhibit color distortions such as green or blue biases caused by wavelength-dependent light absorption in water. Image degradation and distortion due to turbidity become more pronounced as the distance between the target and camera increases. The
level and type of distortion also depend on the type of water in the scene. Although physics-based models offer potential solutions, their limited robustness across different underwater environments challenges their practical implementation.

Recently, data-driven methods enabled by advances in deep learning have demonstrated outstanding results in image enhancement \cite{peng2021u, tang2022autoenhancer, islam2020fast, li2017watergan, sharma2023wavelength, uplavikar2019all}. Furthermore, the inclusion of attention mechanisms has resulted in transformer-based image enhancement methods \cite{peng2021u, tang2022autoenhancer} that exhibit superior performance. They demonstrated that the transformer method, which understands the correlation of neighboring pixels, performs well in the evaluation of restoration. However, the performance of data-driven methods heavily relies on the quality and quantity of training datasets.

As demonstrated in \figref{fig:fig_fist_main_block}, this paper presents a novel approach for \textbf{simultaneous underwater image enhancement and depth estimation} to underwater images, based on an underwater optical model that considers the relationship between scene depth and image degradation. The proposed approach can capture the features and attention of each task during end-to-end training, effectively combining them to improve the quality of the resulting images and depth maps. Additionally, we formulated a \ac{SOTA} single model with an attention mechanism for \textbf{\ac{UIE}} and \textbf{\ac{UMDE}}. To this end, we propose a synthetic dataset that handles the difficult-to-obtain quantity of real underwater datasets and has a satisfactory quality, reflecting various environmental variables (water depth, illumination, etc.). The contribution points are as follows:

\begin{figure}[t] %
    \centering
    \subfigure[Illustration of the networks and sample results]{%
        \includegraphics[width=\columnwidth]{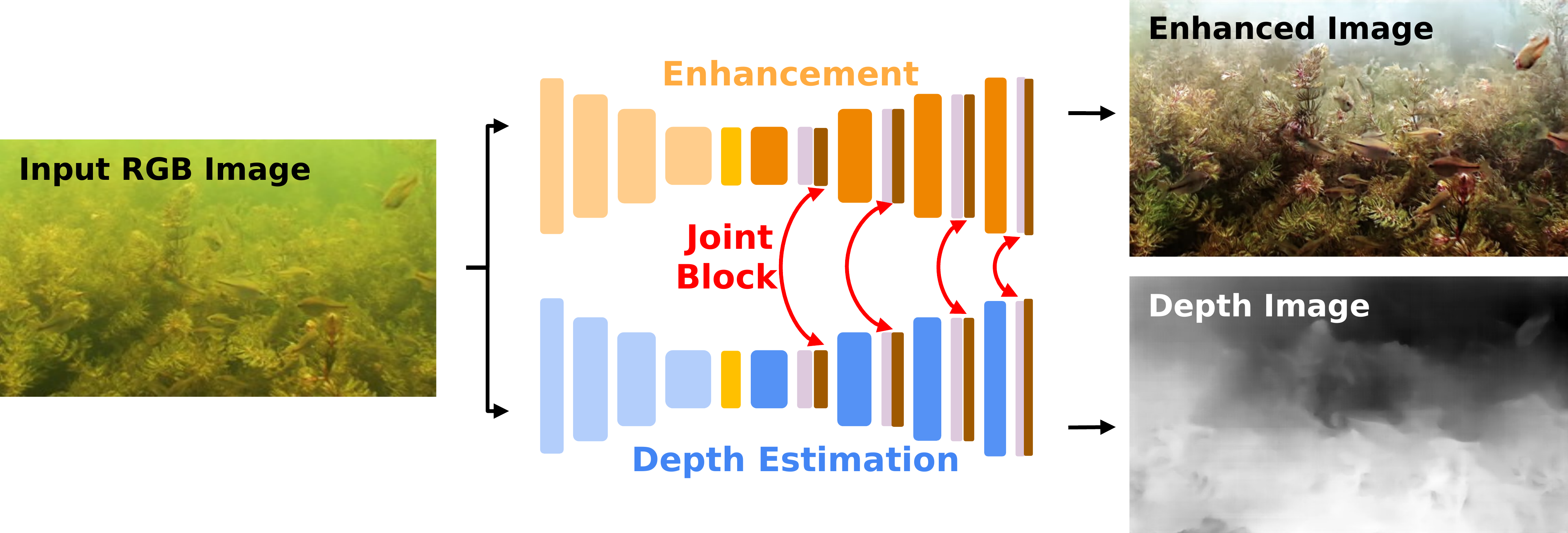}
    } 
    
    \subfigure[Enhancement Comparisons]{%
        \includegraphics[width=\columnwidth]{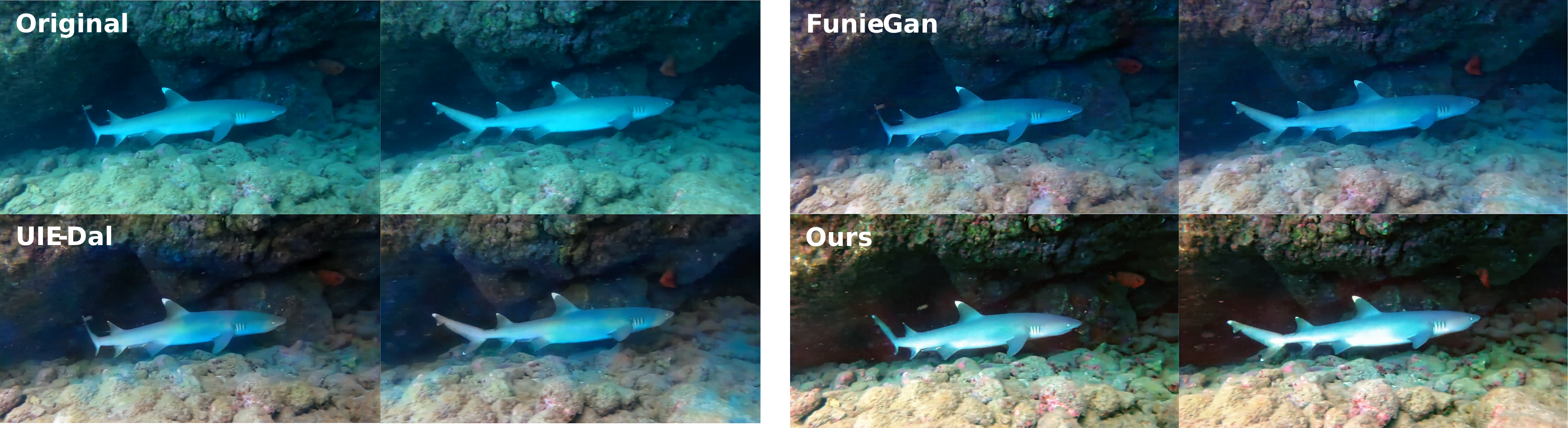}
    } \label{sub_first_fig} 
    \vspace{-3mm}
    \caption{Illustration of the network and sample results. (a) A simplified illustration of the proposed network. A joint block has been constructed between the two models (red arrow). A detailed illustration is shown in \figref{fig:fig_joint_networks}. (b) Enhancement results for consecutive frames. The comparison of feature matching can be seen in \figref{fig:app_matching}.}
    \label{fig:fig_fist_main_block}
\vspace{-7mm}
\end{figure}

\begin{itemize}
    \item We proposed transformer-based U-net architectures for \ac{UIE} and \ac{UMDE}, which show \ac{SOTA} performance for each task.
    \item We designed a novel joint-learning architecture to improve the performance of \ac{UIE} and \ac{UMDE}, which correlate depth and degree of turbidity as prior factors. 
    \item We produced a large-scale synthetic dataset comprising underwater images with dense depth information and from diverse water types.
    \item We demonstrated the robustness of the model for robot vision applications (feature matching and saliency detection). 
\end{itemize}
\section{related works}
\subsection{Underwater Image Enhancement}
Several studies have attempted \ac{UIE} using model, fusion, and learning-based methods that perform robust enhancement through the deep learning.
Model-based methods perform color correction using traditional image processing techniques or by estimating depth based on the amount of blurring. A fusion based algorithm is a method of image restoration by using classical image processing techniques such as histogram equalization and laplacian filter simultaneously with a model-based approach.
Model- and fusion-based methods have been widely studied for restoring and enhancing underwater scenes. 
For instance, \citet{galdran2015automatic} improved contrast by recovering the color associated with short wavelengths.
\citet{berman2017diving} proposed a haze line-based approach to estimate an ambient color and transmission in a color-biased environment.
The retinex theory was also utilized in \cite{fu2014retinex} to improve an image’s contrast or sharpness for underwater dehazing. 
\citet{ancuti2012enhancing} first proposed a fusion-based method that used four weight maps to improve the visibility of far-distance objects deteriorated by medium scattering and absorption. And, \citet{yin2023multiscale} restores colorfulness by performing an equalization that is flexible according to the color space, called color-preserving adaptive histogram equalization.
\citet{cho2018model} proposed a fusion-based method to balance enhancement and detail clearing of images. %

Deep learning has significantly contributed to the \ac{UIE} field as well.
For example, \citet{peng2021u} applied a U-shaped transformer model to \ac{UIE}.
Other studies have directly generated enhanced images using generative models \cite{islam2020fast, li2017watergan}. \citet{islam2020fast} applied a conditional \ac{GAN} to real-time robotic vision, adding an objective function that evaluated the perceptual image quality of the learned image for supervision. %
To correct the color of underwater images in \cite{li2017watergan}, an unsupervised pipeline was furnished with a model that synthesized realistic underwater images by concatenating depth information and non-hazed images. 
To quantitatively and qualitatively improve performance, \citet{sharma2023wavelength} dehazed underwater scenes using an adaptive receptive field in a color channel and applied underwater image semantic segmentation and driver’s 2D pose estimation.  The model by \citet{uplavikar2019all} learned the content features of the water domain to which an image belongs, while distancing from where it does not belong. Object detection has also been used as an evaluation method.

\subsection{Depth Estimation}
Due to its importance in robot vision tasks, many studies have investigated depth estimation, utilizing various approaches such as encoder-decoder structures \cite{dao2022fastmde, yuan2022neural, kim2022global, rahman2023dwinformer, cheng2021swin}. %
\citet{yuan2022neural} used a swin transformer as the encoder and window Fully Connected-CRFs as the decoder, while \cite{kim2022global} used a hierarchical transformer that captures the global context as the encoder with the lightweight decoder.
\citet{rahman2023dwinformer} used two transformers, the dual window self-attention transformer (Dwin-SAT) and the dual window cross attention transformer (Dwin-CAT), which learned pixel-correlation well and performed at the level of SOTA.
\citet{cheng2021swin} predicted depth using a transformer-based encoder-decoder network method through the Multi-Scale Fusion Attention Module.
\citet{valdez2022improving} added semantic segmentation to provide critical information to the depth estimator. In \cite{li2022binsformer}, a classification regression model was formulated in which the regression step directly predicted the continuous depth in pixel units, and the classification step produced probabilistic representations.
\citet{li2022depthformer} introduced an efficient attention mechanism to model a global context. Local planar guidance was used in \cite{lee2019big} to determine whether specific pixels were on the same plane, extending the process to multiple scales to capture both local and global features. Finally, \cite{kumar2020unrectdepthnet} presented a CNN-based architecture that rectifies multiview depth estimation, through epipolar and lens distortion corrections.

\section{Preliminary}
Light attenuation, which is dependent on the wavelength and backscattering, has a greater impact on image degradation in a water medium than under atmospheric conditions. McGlamery’s underwater light propagation model \cite{chiang2011underwater} is commonly used to express distorted underwater images as:

\begin{equation} 
    I_c(x)=J_c(x) \cdot T_\lambda(x)+B_c^{\infty} \cdot\left(1-T_\lambda(x)\right) 
    \label{eq:synthetic_1}
\end{equation}
\begin{equation} 
    I_c(x)=J_c(x) \cdot e^{-\beta_c d(x)}+B_c^{\infty}\left(1-e^{-\beta_c d(x)}\right)
\end{equation}
, where $I_c$ and $J_c$ represent the distorted underwater and clear undistorted images, respectively, $x$ denotes the coordinates of an image pixel, and $B_c$ is the homogeneous ambient light controlled by sunlight or artificial light. The color channels $c$ $\in {r, g, b}$, wavelength $\lambda$ and scene depth $d(x)$ determine transmission ratio $T_\lambda(x) = e^{\beta_c d(x)}$, which indicates how much color intensity is transferred from the clean image. 
This means that we need to estimate both the distortion coefficient ($B_c$ and $\beta_c$) and depth for a complete image restoration.

 Equation \eqref{eq:synthetic_1} shows the dependency of distortion on the attenuation coefficient $\beta_c$ that varies according to the underwater type and background light of the underwater image.
\section{Proposed Method} \label{Proposed Method}

\begin{figure*}[h!]
	\centering
        \vspace{2mm}
	\def\width{0.975\textwidth}%
        \includegraphics[width=\width]{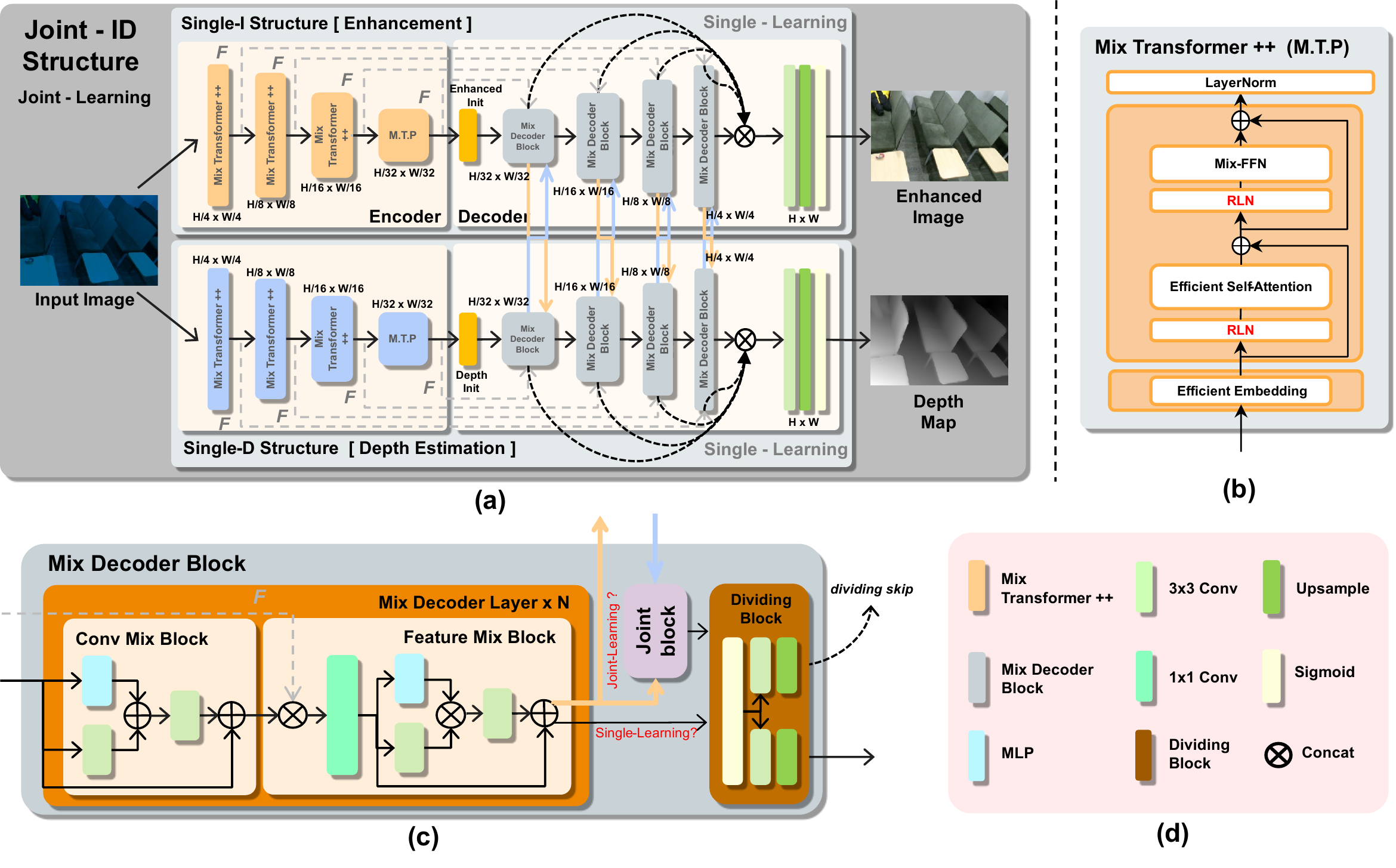}
        \vspace{-1mm}
	\caption{Architecture of the proposed \texttt{Joint-ID} model. (a) The end-to-end architecture consists of two independent models and joint blocks.
 (b) A MixTransformer++ encoder block (c) The Mix Decoder, Joint, and Dividing blocks are applied in three steps: i) after extracting the features through the
encoders, each task in the decoder block is performed, ii) the results of each part are exchanged via the joint block, and iii) the features combined in the block. The feature-sharing process is only used when doing joint learning. (d) Descriptions of blocks and figures.}
	\label{fig:fig_joint_networks}
        \vspace{-1mm}
\end{figure*}

\begin{figure}[h!]
	\centering
        \vspace{2mm}
	\def\width{0.45\textwidth}%
        \includegraphics[width=\width]{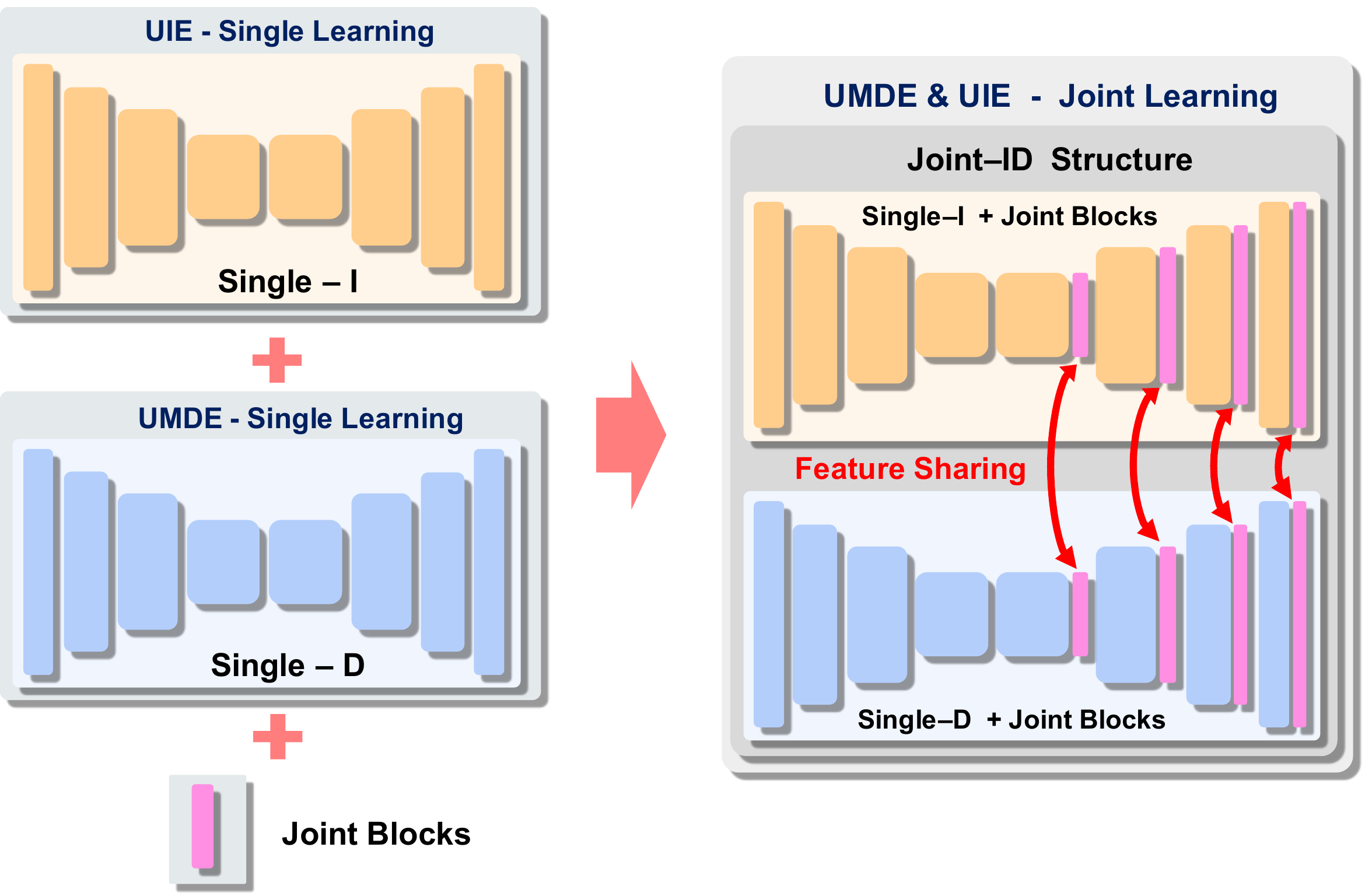}
	\caption{Architecture overview of the Joint-ID Structure. The figure above shows that the Joint-ID Structure is composed of Single-I/D and Joint Block for feature sharing.}
	\label{fig:fig_overview_networks}
        \vspace{-1mm}
\end{figure}

\subsection{Architecture}

As shown in \figref{fig:fig_joint_networks}, the proposed \texttt{Joint-ID} network comprises two separate models, namely, \texttt{Single-I} for \ac{UIE} and \texttt{Single-D} for \ac{UMDE}. These two models use the same U-Net \cite{ronneberger2015u} architecture, which includes four layers of encoder blocks and four layers of decoder blocks, with no joint blocks. The encoder blocks have a feature map of $F_i \in \mathbb{R}^{\frac{H}{2^{i+1}} \times \frac{W}{2^{i+1}} \times C_i}$, where $H$ and $W$ are the height and width of the input image, $C_i$ is the embedding dimension of patches in each layer feature map, and $i = 1,2,3,4$. Understanding local and global features is
crucial to estimate depth and recover an image. Therefore, the encoder resolution starts at $\frac{H}{4} \times \frac{W}{4}$ and decreases to $\frac{H}{32} \times \frac{W}{32}$. The models first extract global features to estimate the absolute depth scale and global hue, as well as local features to estimate depth and recover the image. Subsequently, \texttt{Joint-ID} incorporates joint blocks to perform multi-task learning and enhance the performance of \ac{UMDE} and \ac{UIE}. To understand the overall joint-learning process, an illustration of the architectural overview is shown in \figref{fig:fig_overview_networks}.

\subsection{Mix Transformer++}
In this work, based on Segformer \cite{xie2021segformer}, we propose a modified version of the Mix Transformer block, called \ac{MTP}. The encoder in our model comprises four layers, each having an efficient embedding block. To increase efficiency, we modified each \ac{MTP} layer to consist of $E_i$ = [$E_1, E_2, E_3, E_4$] = $[4, 7, 12, 4]$ blocks from $[3, 4, 18, 4]$. Additionally, the \ac{MTP} block uses a \ac{RLN} \cite{song2022vision} rather than conventional layer normalization, which helps determine the exact scale and bias in the feature maps during the normalization process. Suppose the shape of the feature map is $x_i \in \mathbb{R}^{h_i\times w_i\times c_i}$, where $h_i, w_i$, $c_i$ are the height, width, and number of embedding channels,
respectively. Then, the output of the normalization is:
\begin{equation} 
    \hat{x}_i = \frac{x_i - \mu_i}{\sigma_i}\cdot\gamma_i + \beta_i
    \label{eq:layernormalization}
\end{equation}  
, where $\mu_i$ and $\sigma_i$ are the mean and standard deviation of the values in the feature map, respectively, and $\gamma_i$ and $\beta_i$ are learned scaling factors and biases.  However, unlike \ac{RLN}, the layer normalization that transformer models commonly use is unsuitable for image dehazing \cite{song2022vision}. In \ac{RLN}, the feature map $x_i$, is first input into \eqref{eq:layernormalization} to output $\hat{x}_i$. Then, two linear layers $R_1$, $R_2$, where $\mu, \sigma \in \mathbb{R}^{1\times 1}$, transform the parameters $\mu$ and $\sigma$ via $\gamma, \beta$ to {$W_\gamma$ $\in \mathbb{R}^{1\times c}$ , resulting in:

\begin{eqnarray} 
        \hat{x}^{rln}_i &=& A(\hat{x}_i)\cdot (\sigma_i W_\gamma + B_\gamma) + (\mu_i W_\beta + B_\beta) \nonumber \\
                        &=& A(\hat{x}_i)\cdot R_1(\sigma_i) + R_2(\mu_i)
\end{eqnarray}  
, where $A$ denotes the self-attention efficiency in the Mix Transformer block. $W_\beta$ and $B_\beta$ are weights and bias for the output of the convolution operation $R_2(\mu_i)$, respectively, $W_\gamma$ and $B_\gamma$ are weights and bias for the output of the convolution operation $R_1(\sigma_i)$. $C_i$ = $[C_1, C_2, C_3, C_4] = [64, 128, 320, 512]$ are the dimensions of the feature patches.

\subsection{Mix Decoder}
The decoder network in the proposed model consists of three components: an initialization block, a mix decoder block, and a dividing block. The initialization block comprises two convolutional blocks and serves as a bridge between the encoder and decoder. The mix decoder block has two sections, as shown in \figref{fig:fig_joint_networks}: a convolutional mix block and a feature mix block. Each layer of the decoder consists of a dividing block and a mixed decoder block comprising $D_i$ = [$D_1, D_2, D_3, D_4$] = [3, 3, 3, 3]. The feature map of each layer in the decoder has a size of ${F_i} \in \mathbb{R} ^{{H \over {2^{6-i}}} \times {W \over {2^{6-i}}} \times C_i}$, where $i = {1, 2, 3, 4}$.
The input to the mix decoder block is simultaneously processed by an MLP and a \ac{CNN} designed to capture global and local features, respectively. The feature mix block merges the features from the encoder and the convolutional mix blocks to jointly learn both local and global features. Enhancement and depth estimation tasks necessitate a greater emphasis on the extraction of global features as compared to tasks such as object detection or classification. So, it is designed to extract simultaneously local features and these critical global features to optimize performance in enhancement and depth estimation. Finally, the dividing block receives the features from the feature mix block and splits them into two parts: one for the skip connection and the other as input to the next layer. This is for problem of gradient vanishing from the 1x1 Conv in each Mix Decoder Block, i.e., as the Mix Decoder Block layer is stacked, the local features are trained more intensively, and dividing skip connection is used to solve this problem.

\subsection{Joint Block}

\begin{figure}[h]
	\centering
	\vspace{-1mm}
        \def\width{0.975\columnwidth}%
        \includegraphics[trim = 80 110 40 110, clip, width=\width]{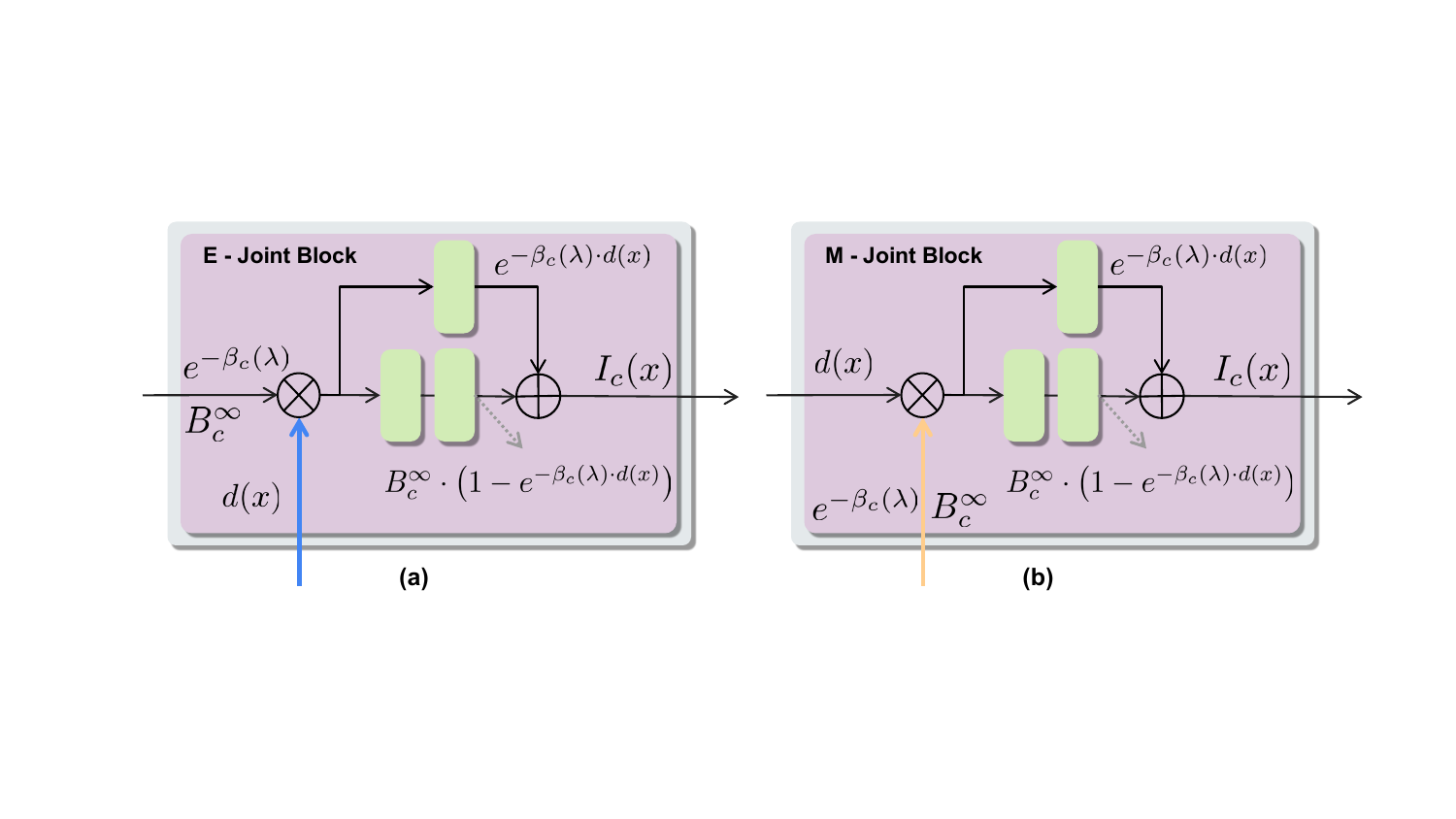}
        \vspace{-1mm}
	\caption{Architecture of the joint block. (a) \texttt{E-joint block} that improves \ac{UIE} by computing $I_c(x)$ from output of previous block and depth information (i.e., with $B_c^{\infty}$, $e^{-\beta_c(\lambda)}$ and depth data $d$). (b) \texttt{M-joint block} that applies the operation opposite to that of the \texttt{E-joint block.} 
 The \texttt{M-joint block} is used to fuse image contrast information and haze features from \texttt{Single-I} with depth d from the previous layer to get Ic(x) so that depth estimation can be performed for different environments.
 } 
	\label{fig:fig_joint_block}
	\vspace{-4mm}
\end{figure}

\begin{figure}[h]
	\centering
        \vspace{2mm}
	\def\width{0.95\columnwidth}%
		\includegraphics[width=\width]{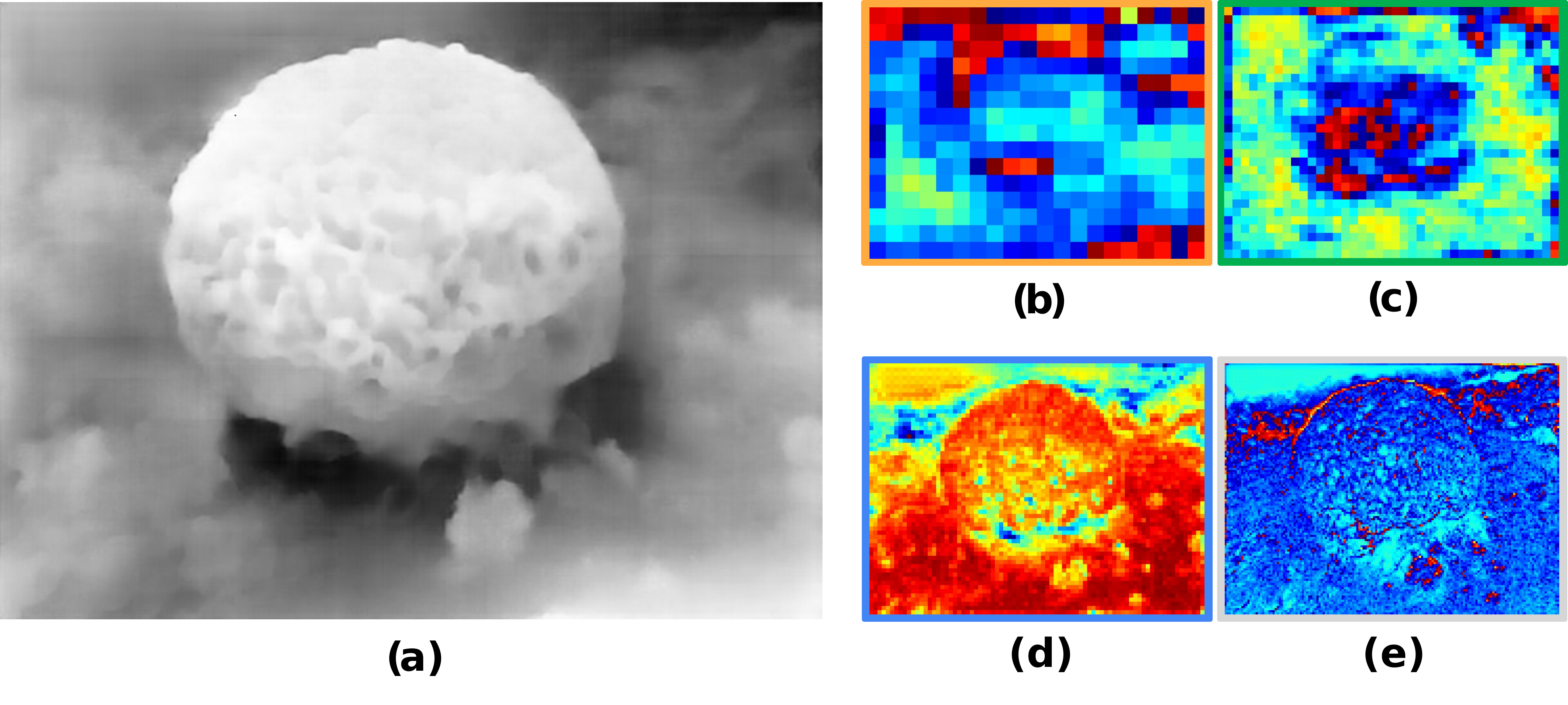}
        \vspace{-3mm}
	\caption{Sample depth and attention maps. (a) Example of depth estimation output. (b), (c), (d), (e) Attention maps of the depth feature in each $i$th \texttt{M-joint block}; maps (b),(c) regard global depth features, whereas maps (d),(e) correspond to local depth features.}
	\label{fig:fig_attnblock}
	\vspace{-2mm}
\end{figure}

Depth and distortion features are extracted from each model by processing with the two separate models in parallel. Subsequently, a joint block combines these two feature sets through a physical propagation model. As shown in \figref{fig:fig_joint_block}, the \texttt{E-joint block}  fuses the enhancement features from the front layer and depth features from the other model, whereas the \texttt{M-joint block} does the opposite. \figref{fig:fig_attnblock}  shows the attention map of the depth information delivered to each \texttt{E-Joint Block}. This process is beneficial for getting the enhanced image in the \texttt{E-joint block} because estimating the depth makes it easier to obtain the distortion coefficients. The block comprises a transmission term that reduces the intensity of the original image and two layers that estimate the distortion due to ambient light. In the $i$th Joint Block, $P^1_i$ is the layer that estimates the transmission term, and $P^2_i$ and $P^3_i$ are the layers that estimate the distortion term. The color recovery process using these layers is as follows:

\vspace{-2mm}
\begin{equation} 
    P^1_i(\lambda, \beta_c, d(x_i)) = e^{-\beta_c(\lambda)\cdot d(x)}
    \label{eq:joint_transmission}
\end{equation} 
\begin{equation} 
    P^3_i(P^2_i(\lambda, \beta_c, d(x_i))) = B_c^{\infty} \cdot\left(1-e^{-\beta_c(\lambda) \cdot d(x)}\right)
    \label{eq:joint_hazaing}
\end{equation}

Based on equations \eqref{eq:joint_transmission} and \eqref{eq:joint_hazaing}, the distorted image can be expressed as:
\begin{equation} 
    I_c(x)=J_c(x) \cdot e^{-\beta_c(x) d(x)}+B_c^{\infty} \left(1-e^{-\beta_c(\lambda)\cdot d(x)}\right)
    \label{eq:joint_output}
\end{equation}

\subsection{Loss Functions}
\textbf{Single UIE Training Loss}. We utilized two types of losses for single enhancement: L2 norm loss, $L_2$,  and Laplacian loss, $L_{Lap}$\cite{li2017laplacian}), are given by the following equations:
\begin{equation} 
    L_2 = \sum_{i,j}{} \sqrt{(\hat{x}_{i,j}-x_{i,j})^2} 
    \label{eq:loss_l2}
\end{equation}

\begin{equation} 
    L_{Lap} = \sum_{i,j}{} |K(\hat{x}_{i,j})-K(x_{i,j})| 
    \label{eq:loss_lap}
\end{equation}
, where $x_{i,j}$ is the ground truth pixel intensity, $i$ and $j$ correspond to the height and width coordinates, respectively, $\hat{x}_{i,j}$ is the network output, and $K$ is the kernel proposed by \citet{li2017laplacian}, which captures edge features. Therefore, using losses \eqref{eq:loss_l2} and \eqref{eq:loss_lap}, we can compute the total loss as:
\begin{equation} 
    L_{enhanced} = k_1(e_1\cdot L_{2} + e_2\cdot L_{Lap})
    \label{eq:loss_enhanced}
\end{equation}
, where $e_1$, $e_2$ and $k_1$ are the weights of the $L_2$, $L_{Lap}$ and total losses, which were set to 1, 0.1, and  to 1.2, respectively.

\textbf{Single UMDE Training Loss}. \ac{UMDE} used only one kind of loss called silog $S(g)$ \cite{eigen2014depth}, which is given by: 

\begin{equation} 
    S(g) = \frac{1}{T}\sum_{i}g_i^2 - \frac{e_3}{T^2}\left(\sum_{i}g_i^2\right)^2
    \label{eq:depth_loss_1}
\end{equation}
, where $g_i = \log\hat{d_i} - \log{d_i}$, $d_i$ is ground truth depth, and $e_3$ is the loss weight which we set to 0.85. \citet{lee2019big}, proposed an enhanced silog loss, as follows:

\begin{equation} 
    L_{depth} = k_2\sqrt{S(g)} ,
    \label{eq:depth_loss_2}
\end{equation}
, where $k_2$ is a constant we set to 1.0. In this study, we used the improved $L_{depth}$ loss.

\textbf{Joint-learning with UIE and UMDE Training}. When performing joint learning, we use \eqref{eq:loss_enhanced} and \eqref{eq:depth_loss_2}, varying only constraints k1 and k2. Therefore, the
final UIE-UMDE joint loss is: 
\begin{equation} 
    L_{joint} = k_1(e_1\cdot L_{2} + e_2\cdot L_{Lap}) + k_2\sqrt{S(g)}
    \label{eq:joint_loss}
\end{equation}
, where $k_1$ is 2.0, and $k_2$ is 1.5.

\subsection{Dataset preparation} \label{Dataset preparation}

\begin{figure}[!h]
    \def\width{1\columnwidth}%
        \includegraphics[width=\width]{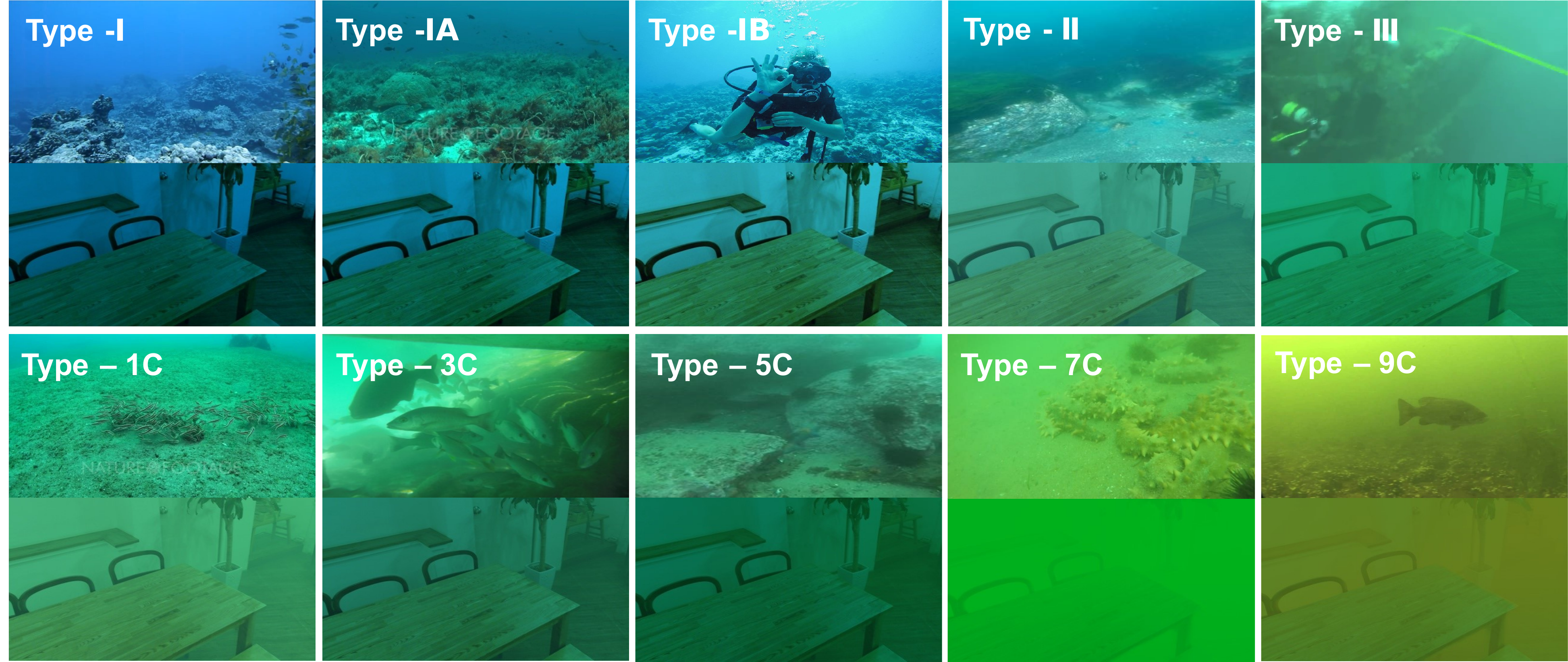}
        \vspace{-3mm}
    \caption{Sample images in our dataset. The dataset includes 10 water types according to \cite{solonenko2015inherent}, (i.e, there are '\uppercase\expandafter{\romannumeral1}', '\uppercase\expandafter{\romannumeral1}A', '\uppercase\expandafter{\romannumeral1}B', \uppercase\expandafter{\romannumeral2}, '\uppercase\expandafter{\romannumeral3}, 1C, 3C, 5C, 7C and 9C water types.)}
    \label{fig:fig_dataset_samples}
    \vspace{-1mm}
\end{figure}

According to \citet{solonenko2015inherent}, the attenuation coefficient $\beta_c$ is defined as sum of absorption coefficient $a_c$, scattering coefficient $b_c$ (i.e., $\beta_c$ = $a_c$ + $b_c$). So, using this theory, \citet{wang2021single} introduce that ambient light coefficient can be modeled as follows:

\begin{equation} 
    \begin{split}
    B_c^{\infty}(\lambda) 
                          &= \frac{b(\lambda)\cdot e^{-[a(\lambda) + b(\lambda)]\cdot D}}{a(\lambda) + b(\lambda)}\cdot E(0, \lambda),
    \end{split}
    \label{eq:synthetic_3}
\end{equation}  

 where $D$ is surface-object distance and $E(0, \lambda)$ = $E^{in}_c(x)$ ($E(0, \lambda)$ is the ambient light at the sea surface). Also, \citet{chiang2011underwater} says that $E(0, \lambda)$ has the same value for each channels $c$ ($c$ $\in$ [$r$, $g$, $b$]) above the water surface. So, the equation about ambient light coefficient finally can be written down as below:

\begin{equation} 
    \frac{B_r}{B_g} =  \frac{b_r}{b_g}\cdot \frac{a_g + b_g}{a_r + b_r}\cdot e^{-[a_r+b_r-a_g-b_g]\cdot D}
    \label{eq:synthetic_3}
\end{equation}  

Similarly,
\begin{equation} 
    \frac{B_b}{B_g} =  \frac{b_b}{b_g}\cdot \frac{a_g + b_g}{a_b + b_b}\cdot e^{-[a_b+b_b-a_g-b_g]\cdot D},
    \label{eq:synthetic_3}
\end{equation}  

Finally, the above equation defines the relation between $B_g$, $B_r$, and $B_b$. Therefore, if $B_g$ is determined randomly, $B_r$ and $B_b$ may be obtained. In this paper, we adopt the algorithm in \cite{ueda2019underwater} and \cite{wang2021single}. The details of the process, $B_g$, the ambient light coefficients were set to [0.4, 0.5, 0.6, 0.7], and the absorption coefficient $a$ and scattering coefficient $b$ were assigned as in \cite{solonenko2015inherent}. We tried to form a synthetic dataset with depth $D$ $\in [1.5, 3, 5, 7.5, 10, 15]$ according to the water types in \cite{solonenko2015inherent}. For water type '\uppercase\expandafter{\romannumeral1}', '\uppercase\expandafter{\romannumeral1}A', '\uppercase\expandafter{\romannumeral1}B', the depth was [3, 5, 7.5, 10, 15]. For type \uppercase\expandafter{\romannumeral2}, depth was [3, 5, 7.5, 10]. For '\uppercase\expandafter{\romannumeral3} and '\uppercase\expandafter{\romannumeral1}C types', depth was [3, 5, 7.5]. For '3C' and '5C', the depth was [3, 5]. Finally, for '9C', we was [1.5, 3]. The ground dataset used was a dataset with an object detail called diml \cite{cho2021diml}.  A total of 48000 synthetic data were generated, of which 47000 were selected for the training set and 1000 for the test set. \figref{fig:fig_dataset_samples} shows sample results from our synthetic dataset.

\section{Experimental Results}
In this section, we first describe training details and basic settings used for the implementation of our proposed \texttt{Joint-ID}. We perform qualitative and quantitative image enhancement and depth estimation experiments on several datasets, including synthetic and real underwater scenes, and compare them with other methods, demonstrating the \ac{SOTA} performance of our model for the two underwater image processing tasks. And, we then conduct an ablation study, which demonstrates its effectiveness in learning depth information and distortion coefficients.

\subsection{Experimental Details}
As mentioned in \ref{Proposed Method}, we utilized by paired data ({$X$}, {$Y$}) from \ref{Dataset preparation} to train the proposed networks for predictive functions $f_1$ : $X \rightarrow Y^{E}, Y^{D}$ , $f_2$ : $X \rightarrow Y^E$, $f_3$ : $X \rightarrow Y^D$, where $f_1$ is function for multi-task learning, $f_2$ is for \ac{UIE}, and $f_3$ is for \ac{UMDE}. In addition, we set the learning rate to $3e^{-4}$ to train the $f_1$ network, and to $2e^{-4}$ for networks $f_2$, $f_3$. The algorithm by Adamw \cite{loshchilovdecoupled} is used for optimization with a mini-batch size of 4, and the default values of $\beta_1$, $\beta_2$ and weight decay are set as 0.9, 0.999 and 1e-8, respectively.

 We conducted several experiments on synthetic and real-world underwater datasets \cite{islam2020fast}, \cite{peng2021u}, \cite{li2019underwater}, \cite{li2017watergan}, \cite{berman2020underwater}, after retraining the other models on the datasets to verify their implementation in the same environment. All the hyperparameters in the training process were the same as those provided by each model. The training and validation were conducted on Python and Pytorch frameworks using NVIDIA RTX3090ti on Ubuntu 22.04.

\subsection{Image Enhancement}
\begin{figure*}[t!] %
    \begin{center}
        \includegraphics[width=2.0\columnwidth]{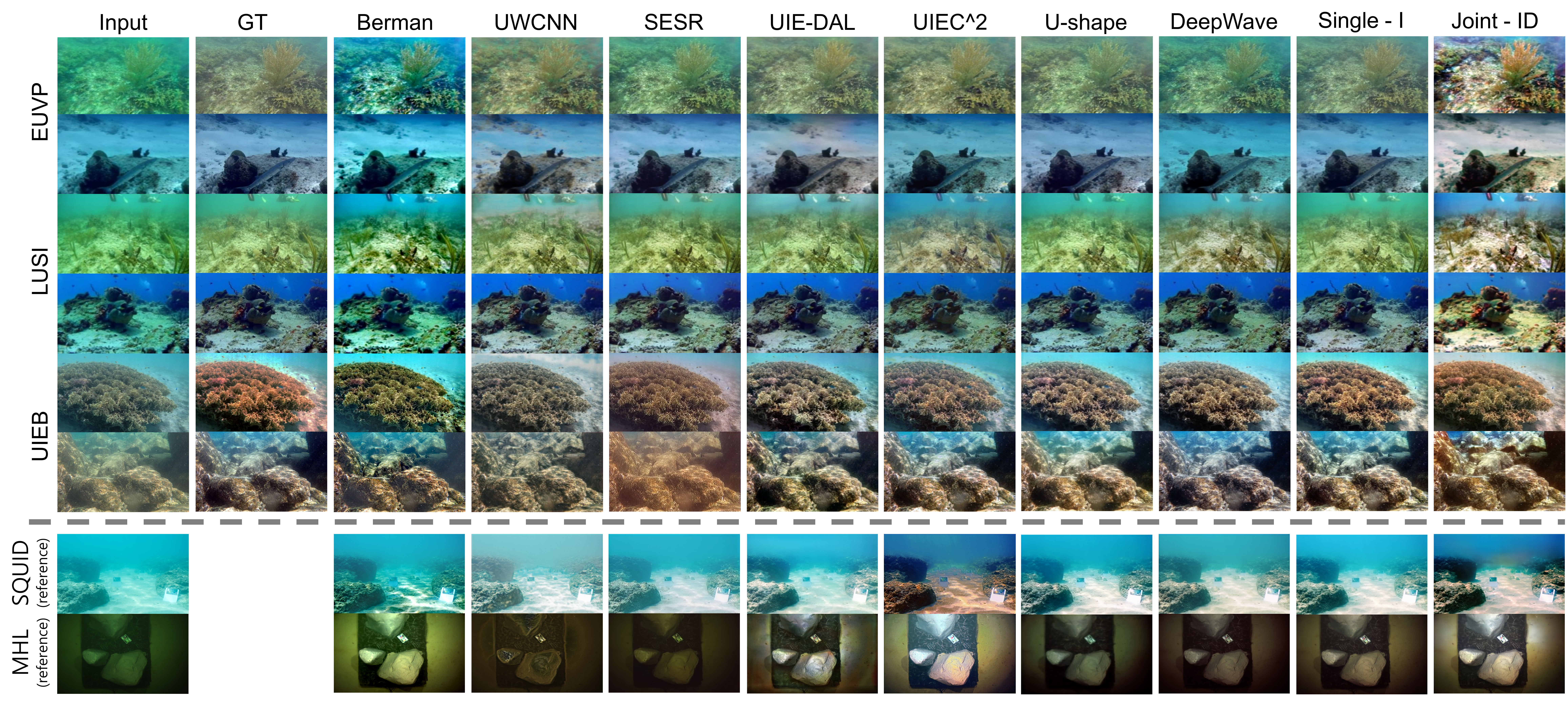}
        \vspace{-2mm}
        \caption{Qualitative evaluation on a reference test set. Results in each row differ in training data and test set; 1) The first and second rows were trained with EUVP and used EUVP's test set. 2) The third and fourth rows were trained with LUSI and used LUSI's test set. 3) The fifth and sixth rows were trained with uieb and used uieb's test set. Finally, the seventh and eighth rows are the result from real underwater datasets(SQUID, MHL) after training with the reference weights of each model.}
        \vspace{-3mm}
        \label{fig:enhancement_GT}
    \end{center}
\end{figure*}

\begin{table*}[t!]
\caption{Quantitative performance comparison for enhancement on PSNR, Sq Rel, SSIM, and UCIQE in our reference dataset and real underwater dataset: the first and second best scores (in each row) are colored red, and blue, respectively}
\label{tab:quantitative_ref2}
\centering
\resizebox{0.95\textwidth}{!}{%
\begin{tabular}{c||c|c|c|c|c|c|c|c|c|c|c|c}
\hline
\multirow{14}{*}{A} &\textbf{Metrics} & \textbf{Dataset} & Berman & UWCNN  & UGAN   & UIEC\textasciicircum{}2 & FUnIE  & UIE-DAL & U-Shape & DeepWave   & \textbf{Single-I} & \textbf{Joint-ID} \\ \cline{2-13}
&                        \multirow{3}{*}{PSNR}   & EUVP             & 17.57  & 24.29  & 23.49  & 24.49   & 23.41  & 24.93   & \textcolor{blue}{26.02}   & 24.01    & \textcolor{red}{26.13}  & 17.5   \\
&                                                & LUSI             & 16.92  & 23.19  & 26.38  & 25.34   & 24.07  & 25.82   & \textcolor{blue}{26.84}   & 24.99    & \textcolor{red}{28.42}  & 18.91  \\ 
&                                                & UIEB             & 16.91  & 19.18  & 19.62  & 19.39   & 19.08  & \textcolor{blue}{20.25}   & 19.7    & 19.67    & \textcolor{red}{20.26}  & 19.31  \\ \cline{2-13}
&                        \multirow{3}{*}{Sq Rel} & EUVP             & 10.12  & 11.052 & 9.937  & 10.925  & 11.353 & 11.042  & \textcolor{red}{8.038}   & 8.484    & \textcolor{blue}{8.336}  & 33.94  \\
&                                                & LUSI             & 3.3    & 5.218  & 3.456  & 14.13   & 4.39   & 3.498   & \textcolor{red}{2.477}   & 3.701    & \textcolor{blue}{3.088}  & 15.72  \\  
&                                                & UIEB             & 20.7   & 65.09  & \textcolor{blue}{14.94}  & 30.49                  & 24.27  & 15.48   & 15.53   & 15.82 & \textcolor{red}{11.5}  & 17.49 \\ \cline{2-13}
&                        \multirow{3}{*}{SSIM}   & EUVP             & 0.58   & 0.695  & 0.681  & 0.701  & 0.664  & \textcolor{blue}{0.728}   & 0.719   & 0.712    & \textcolor{red}{0.734}  & 0.659  \\
&                                                & LUSI             & 0.65   & 0.742  & 0.798  & 0.782                   & 0.773  & \textcolor{blue}{0.801}   & 0.782   & 0.792    & \textcolor{red}{0.814}  & 0.724  \\ 
&                                                & UIEB             & 0.72   & 0.688  & 0.777  & 0.755              & 0.741  & 0.798   & 0.747   & \textcolor{blue}{0.787} & \textcolor{red}{0.811}  & 0.768  \\ \cline{2-13}
&                        \multirow{3}{*}{UCIQE}  & EUVP             & \textcolor{blue}{0.619}  & 0.5764 & 0.581  & 0.5747                  & 0.59   & 0.578   & 0.5722  & 0.5706   & 0.578  & \textcolor{red}{0.6266} \\
&                                                & LUSI             & \textcolor{red}{0.615}  & 0.5527 & 0.5807 & 0.57495                 & 0.5742 & 0.5721  & 0.5808  & 0.5786   & 0.5783 & \textcolor{blue}{0.6122} \\ 
&                                                & UIEB             & 0.614  & 0.547  & 0.61   & 0.609 & 0.599  & 0.596   & 0.607   & 0.598    & \textcolor{red}{0.619}  & \textcolor{blue}{0.617}  \\ \hline

\multirow{4}{*}{B}  &\multirow{2}{*}{\begin{tabular}[c]{@{}c@{}}UCIQE\\ (pretrained)\end{tabular}} & MHL & 0.587  & 0.473 & 0.451 & \textcolor{red}{0.623} & 0.454 & 0.562   & 0.458   & 0.4818   & 0.5802 & \textcolor{blue}{0.592} \\  
                    & & SQUID            & 0.488  & 0.471 & 0.428 & \textcolor{red}{0.613}                   & 0.424  & 0.468   & 0.504   & 0.494    & 0.506  & \textcolor{blue}{0.574} \\ \cline{2-13}
& \begin{tabular}[c]{@{}c@{}} UCIQE\\ (retrained)\end{tabular}                   & SQUID            & 0.488  & 0.313 & 0.463 & \textcolor{blue}{0.552}                   & 0.344  & 0.454   & 0.537   & 0.547    & 0.506  & \textcolor{red}{0.574} \\ \hline
\end{tabular}
}
\end{table*}

We evaluated the proposed joint model (\texttt{Joint-ID}), including our single model (\texttt{Single-I}), with model-based method (Berman\cite{berman2017diving}) and deep learning-based methods (UWCNN\cite{anwar2018deep}, UIE-DAL\cite{uplavikar2019all}, SESR\cite{islam2020simultaneous}, UIEC\^{}2-Net\cite{wang2021uiec}, U-shape\cite{peng2021u},  DeepWave\cite{sharma2023wavelength}) for image enhancement task. We conducted the experiments separately in two sections: (A) The datasets with GT (EUVP\cite{islam2020fast}, LUSI\cite{peng2021u}, UIEB\cite{li2019underwater}) and (B) The datasets without the GT (MHL\cite{li2017watergan}, SQUID\cite{berman2020underwater}). For fair comparisons, we utilized supervised metrics (\ac{PSNR}, \ac{SSIM}, Squared Relative Error (Sq Rel)) and the unsupervised metric (Underwater Color Image Quality Evaluation (UCIQE) \cite{yang2015underwater}).

\textbf{Evaluation for datasets with GT.}
For comparison in supervised metrics, all learning-based methods, excepting (\texttt{Joint-ID}), were trained on standard datasets (EUVP, LUSI, and UIEB), whereas the joint model training used our synthetic dataset. In comparison to unsupervised metrics, we used the training pipeline presented in each paper for each model. The qualitative comparisons are shown in Table A in Table \ref{tab:quantitative_ref2}; moreover, the quantitative evaluation is summarized in \figref{fig:enhancement_GT}'s first to sixth rows.

As \figref{fig:enhancement_GT} demonstrates, Berman method \cite{berman2017diving} fall short in hue rectification, while the other models, with the exception of UWCNN and SESR, generally approximate the color of the ground truth well; UWCNN and SESR show good similarity to ground truth on EUVP and LUSI, but not on UIEB. In comparison, \texttt{Joint-ID}, trained in a different way than the others, has a large color intensity difference from the ground truth.
However, the qualitative results show the method-based multi-task has better color contrast and colorfulness.
For the quantitative comparisons, Table A in \ref{tab:quantitative_ref2} shows that \texttt{Single-I} outperforms other models in terms of PSNR, SSIM and achieves often competitive performance in comparison to Sq Rel, whereas in terms of unsupervised metric (UCIQE), it achieves comparable color recovery. This represents the perceptual enhancement performance of the proposed single network design.

\figref{fig:enhancement_GT} and Table \ref{tab:quantitative_ref2} show that \texttt{Joint-ID} is not qualitatively similar to GT and doesn't score higher on the supervised metrics, respectively. However, we can see that qualitatively GT has a slight color bias and hazing, while \texttt{Joint-ID}-generated images are considerably sharper, achieving comparable color recovery and hue rectification. It indicates that our model has more of a qualitative advantage in underwater robot vision where high color contrast is required. This is also demonstrated quantitatively in Table \ref{tab:quantitative_ref2}. The UCIQE comparison in Table \ref{tab:quantitative_ref2}, evaluating image qualities based on colorfulness, sharpness, and contrast, shows that the joint model outperformed the other models. Combining the experiments takes conclusions: i) \texttt{Single-I} is more potent in supervised metrics. ii) The multi-task model has more qualitative robustness than the model for only an enhancement task, and the qualitative results show ours has better color contrast and colorfulness.

\textbf{Evaluation for datasets without GT.}
To evaluate MHL \cite{li2017watergan} and SQUID \cite{berman2020underwater}, we prepared two types of trained weights per model, excluding ours, one for the weight given by the model, and the other for the weight retrained using our synthetic dataset. The reason for preparing the retrained weights is to provide a rigorous evaluation. %
As seen in the seventh to eighth rows in \figref{fig:enhancement_GT}, for the SQUID \cite{berman2020underwater} datasets, all methods except for UIEC\^{}2 and our joint model, did not effectively remove the blue color bias and hazing from the entire scene. These tendencies are seen similarly in the MHL datasets, which have harsher turbidity. In MHL comparison, UIEC\^{}2 \cite{wang2021uiec}, UIE-DAL \cite{uplavikar2019all} and our joint model estimate backscattering coefficient well, while other methods generally suffer from dark scene. Also, the results of UIE-DAL \cite{uplavikar2019all} have some artifacts, but our proposed joint model shows no biased color- or edge-preserving ability for objects such as stones. Based on the qualitative results, in \ref{tab:quantitative_ref2}, table B's first and second row also show UICE\^{}2 and our joint model has good results for metric UCIQE. 
However, the models are strictly not trained using the same dataset (i.e., one is on the UIEB and the other on our dataset), which prevents us from comprehending which model is more robust in the evaluation.
The third in Table B shows that our model is best on the metric UCIQE, and it shows that our model has much-improved restoration performance in cases with severe turbidity and some biased water domains. %

\subsection{Depth Estimation}

\begin{figure*}[t!]
	\centering
	\def\width{0.95\textwidth}%
	\includegraphics[width=\width]{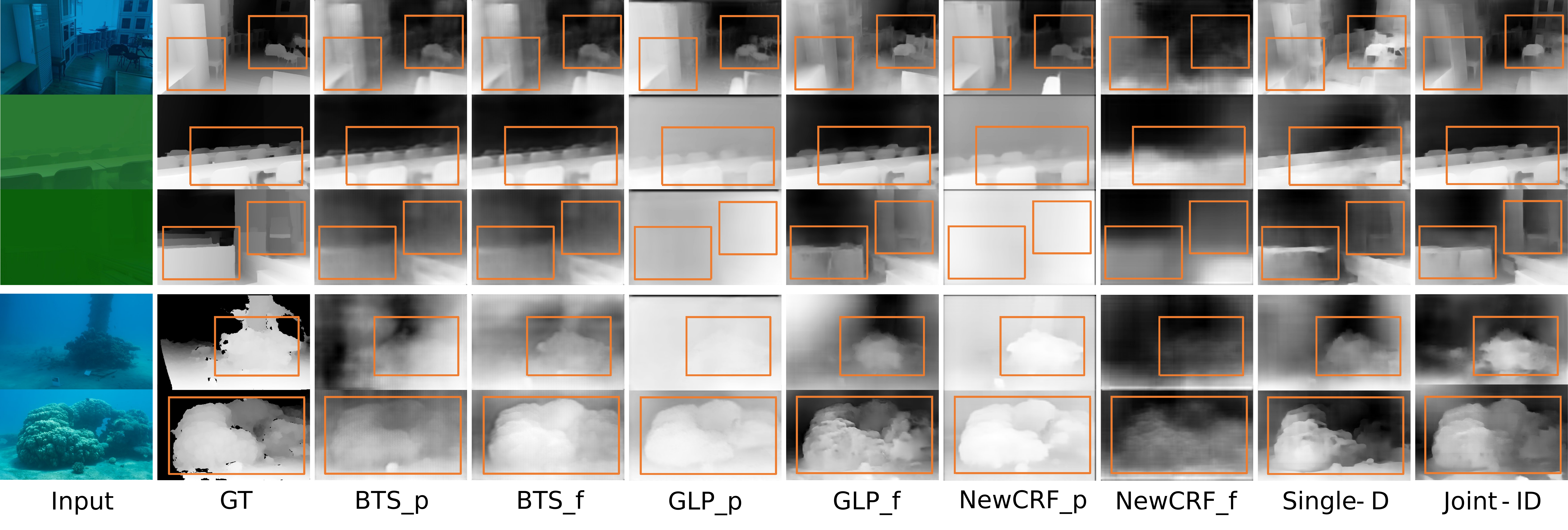}
	\vspace{-2mm}
        \caption{Qualitative comparison with other models in \ac{UMDE}. The evaluation runs on a synthetic dataset and SQUID. Our method achieves more distinctive results, especially on object boundaries (orange box).}
        \label{fig:depth_estimation_SQUID}
        \vspace{1mm}
\end{figure*}

\begin{table*}[h]
\vspace{-3mm}
\caption{Quantitative Evaluation for Synthetic Dataset and SQUID Dataset: the first and second best scores (in each row) are colored red, and blue, respectively}
\vspace{-1mm}
\label{tab:depth_estimation_synthetic}
\centering
\resizebox{0.8\textwidth}{!}{
\begin{tabular}{c||c|c|c|c|c|c|c|c}
\hline
Test set &    Method                & silog $\downarrow$ & Abs Rel $\downarrow$ & RMS Err $\downarrow$   & Sq Rel $\downarrow$ & d1 $\uparrow$ & d2 $\uparrow$   & d3 $\uparrow$    \\ \hline
\multirow{8}{*}{Synthetic} & BTS\_p      & 21.74 & 0.26    & 1.04 & 0.28   & 0.63 & 0.87 & 0.92 \\
               & BTS\_f       & 18.47 & \textcolor{blue}{0.21}    & 0.64 & \textcolor{red}{0.18}   & 0.64 & 0.90 & 0.94 \\
               & GLPDepth\_p & 19.04 & 0.49    & 3.88 & 2.55   & 0.33 & 0.66 & 0.87 \\
               & GLPDepth\_f  & 17.30 & 0.23    & 2.02 & 0.79   & \textcolor{blue}{0.67} & \textcolor{blue}{0.90} & 0.96 \\
               & NewCRF\_p   & 18.72 & 0.47    & 3.82 & 2.31   & 0.31 & 0.65 & 0.88 \\
               & NewCRF\_f    & 22.29 & 0.26    & 2.43 & 0.92   & 0.57 & 0.86 & 0.95 \\
               & \textbf{Single-D}      & \textcolor{blue}{16.52} & 0.23    & \textcolor{blue}{0.68} & \textcolor{blue}{0.23}   & 0.63 & 0.88 & \textcolor{blue}{0.96} \\
               & \textbf{Joint-ID}      & \textcolor{red}{16.08} & \textcolor{red}{0.21}    & \textcolor{red}{0.62} & 0.65  & \textcolor{red}{0.68}  & \textcolor{red}{0.91} & \textcolor{red}{0.97} \\ \hline
\multirow{8}{*}{SQUID}  & BTS\_p      & 16.24 & 0.45    & \textcolor{blue}{1.78} & 0.87   & 0.18 & 0.62 & \textcolor{red}{1.00} \\
               & BTS\_f       & 16.34 & 0.27    & 1.39 & 0.40   & 0.22 & 0.78 & 0.95 \\
               & GLPDepth\_p & 19.04 & 0.49    & 3.88 & 2.55   & 0.33 & 0.66 & 0.87 \\
               & GLPDepth\_f  & 17.30 & 0.23    & 2.02 & 0.79   & 0.67 & 0.90 & 0.96 \\
               & NewCRF\_p   & 18.71 & 0.47    & 3.82 & 2.31   & 0.31 & 0.65 & 0.88 \\
               & NewCRF\_f    & 22.29 & 0.26    & 2.43 & 0.92   & 0.57 & 0.86 & 0.95 \\
               & \textbf{Single-D}      & \textcolor{blue}{14.21} & \textcolor{blue}{0.19}  & \textcolor{red}{0.92} & \textcolor{blue}{0.22} & \textcolor{blue}{0.82} & \textcolor{blue}{0.97} & \textcolor{red}{1.00} \\
               & \textbf{Joint-ID}      & \textcolor{red}{12.41} & \textcolor{red}{0.09}    & 1.805 & \textcolor{red}{0.21}   & \textcolor{red}{0.87} & \textcolor{red}{1.00} & \textcolor{red}{1.00} \\ \hline
\end{tabular}
}
\vspace{-1mm}
\end{table*}

For depth estimation, we evaluated our model qualitatively and quantitatively with other ground scene-targeted deep learning-based monocular depth estimation models (BTS \cite{lee2019big}, GLPDepth \cite{kim2022global}, NewCRF \cite{yuan2022neural}). For a fair evaluation of underwater scenes, we utilized both a pretrained model and a fine-tuned model for underwater synthetic scenes while testing the ground depth estimation models. The pretrained model for ground depth estimation is named p, and the fine-tuned model is named f. We tested our synthetic dataset for various underwater environmental conditions and some samples of the SQUID dataset for real underwater scenes. Quantitative evaluation of the SQUID datasets was performed on an object-centric image with a limited depth scale. We utilized well-known supervised metrics (silog, Absolute Relative Error (Abs Rel), Route Mean Square Error (RMS Err) Accuracy under threshold (d1, d2, d3)) \cite{lee2019big}.

\textbf{Evaluation for Synthetic Datasets.}
In the case of the low-turbidity first row in \figref{fig:depth_estimation_SQUID}, almost all methods succeeded in estimating the depth almost similar to GT, except for the NewCRF\_f model. In contrast, when the turbidity was high (i.e., two and three rows), both GLP\_p and NewCRF\_p failed to estimate the depth. Both BTS models roughly estimated the shape of an object; however, the edges were unclear.  The GLP\_f performed similarly to the GT depth map, even under extreme conditions. Also, \texttt{Joint-ID} showed good performance with high turbidity and made the contrast of the edge more identifiable. Table \ref{tab:depth_estimation_synthetic} shows that our joint model has a powerful performance similar to that of \ac{SOTA} in terms of the metrics (silog, Abs rel, RMS Err, d1, d2, d3).

\textbf{Evaluation for SQUID Datasets.}
Both modes of BTS, NewCRF, and GLP\_p failed to estimate the overall depth-scale image. GLP\_f estimated the depth scale and shape of objects well, and our single model followed the GLP\_f performance. %
However, it is precarious to estimate the depth of the background according to the first-row image. On the other hand, our joint model shows accurate performance over a wider depth scale with relatively better background depth estimation.
Table \ref{tab:depth_estimation_synthetic} show that our joint model performs almost well as \ac{SOTA} in the metrics(silog, Abs Rel, Sq Rel, d1, d2, d3). 

Overall, the single model performs better than the existing methodologies, and the joint model performs the most powerful performance. This shows that feature sharing through joint learning plays a crucial role in improving the performance of depth estimation.

\subsection{Ablation Study}
We analyzed the comparison results between our single and joint models for the \ac{UIE} and \ac{UMDE}, both quantitatively and qualitatively. An ablation study of multi-task learning was conducted for the synthetic dataset, representing diverse water types.

\begin{figure}[h!]
    \begin{center}
        \vspace{-2mm}
        \def\width{1\columnwidth}%
            \includegraphics[width=\width]{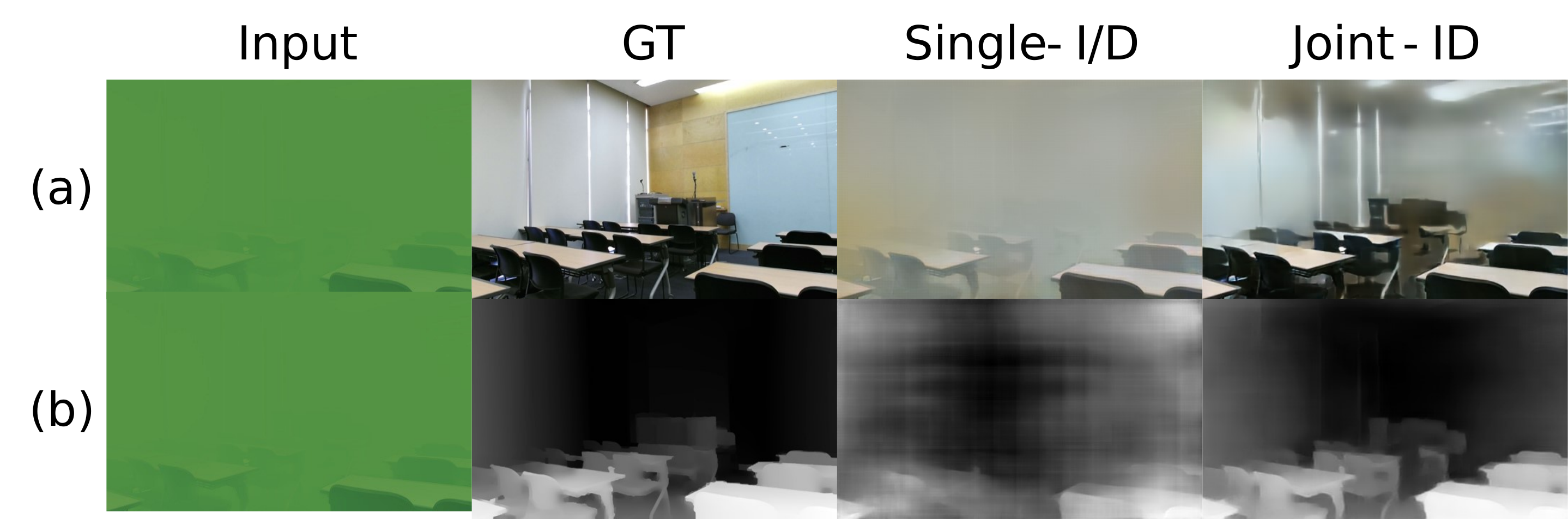}
            \vspace{-4mm}
            \caption{Qualitative \ac{UIE} and \ac{UMDE} results from our single models and Joint-ID on a few synthetic data samples. (a) Qualitative results from input sample with weak turbidity and strong distortion for distance in \ac{UIE}. (b) Qualitative results from a sample in \ac{UMDE}.}
            \label{fig:ablation_study}
     \end{center}
\vspace{-1mm}
\end{figure}

 In the \figref{fig:ablation_study}, the input images show a difficult case: the single model only estimates nearby objects, but it does not include a deeper scale. A single model may be vulnerable to high turbidity and wide depth scales; however, our joint model shows for our joint model, not only each nearby object but also the depth scale up to the background was accurately estimated. Our experiments demonstrate that using joint learning, both depth and distortion information help each other and derive good results.

\begin{table}[h!]
    \vspace{-1mm}
    \centering
    \caption{Quantitative performance comparison for enhancement on our synthetic dataset}
    \vspace{-1mm}
    \label{tab:image_enhancement}
    \resizebox{0.9\columnwidth}{!}{
    \begin{tabular}{c||cccc}
\toprule
Method           & PSNR $\uparrow$  & SSIM $\uparrow$  & Abs Rel $\downarrow$ & Sq Rel $\downarrow$ \\ \midrule
Single - I  & 24.45 & 0.658 & 3.47     & 0.803   \\ 
Joint - ID  & $\boldsymbol{25.85}$ & $\boldsymbol{0.728}$ & $\boldsymbol{1.60}$     & $\boldsymbol{0.270}$   \\ \bottomrule
\end{tabular}
}
\vspace{-2mm}
\end{table}

\begin{table}[h!]
    \vspace{-2mm}
    \centering
    \caption{Quantitative performance comparison for depth estimation on our synthetic dataset}
    \vspace{-2mm}
    \label{tab:depth_estimation}
    \resizebox{0.95\columnwidth}{!}{
    \begin{tabular}{c||cccccc}
    \cmidrule(r){1-7}
    Method           & Abs Rel $\downarrow$ & Sq Rel $\downarrow$ & RMS Err $\downarrow$ & d1 $\uparrow$   & d2 $\uparrow $   & d3 $\uparrow$    \\ \cmidrule(r){1-7}
    Single - D & 0.231    & 0.239   & 0.686    & 0.637 & 0.883 & 0.966 \\
    Joint - ID & $\boldsymbol{0.211}$    & $\boldsymbol{0.195}$   & $\boldsymbol{0.626}$    & $\boldsymbol{0.67}$  & $\boldsymbol{0.912}$ & $\boldsymbol{0.975}$ \\ \cmidrule(r){1-7}
    \end{tabular}
    }
\vspace{-1mm}
\end{table}

The fact that our proposed joint model is superior to a single model is proved quantitatively on both two tasks, \ac{UIE} and \ac{UMDE}. In Table \ref{tab:depth_estimation}, the metrics(Abs Rel, Sq Rel, RMS Err, d1, d2, d3) indicate that the joint model estimates a wider depth scale than a single model. In addition, as \ref{tab:image_enhancement} illustrates, the metrics(PSNR, SSIM, Sq Rel) indicate that the joint model performs better than a single model for image enhancement.

\section{Application to robot vision} \label{application}
\subsection{Feature Matching}
\figref{fig:app_matching} shows the matching of ORB features in a frame of a publicly available underwater video, as well as a plot of the number of inlier features in all the frames of the video. It is evident from \figref{fig:app_matching}(b) that all \ac{UIE} methods capture more feature matching pairs than the original input images, with our model having particularly more ORB features than the other \ac{UIE} models. This indicates that our model has superior contrast and dehazing abilities than the other methods.

As shown in \figref{fig:app_matching}(a), the original underwater image exhibits very few ORB features, leading to potential drift errors. In contrast, the ability of our model to restore color and contrast produces rich features, enabling stable odometry performance.

\begin{figure}[h!] %
    \centering
    \vspace{-3mm}
    \subfigure[Feature Matching (top: Original Image, bottom: Proposed Image)]{%
        \includegraphics[width=0.8\columnwidth]{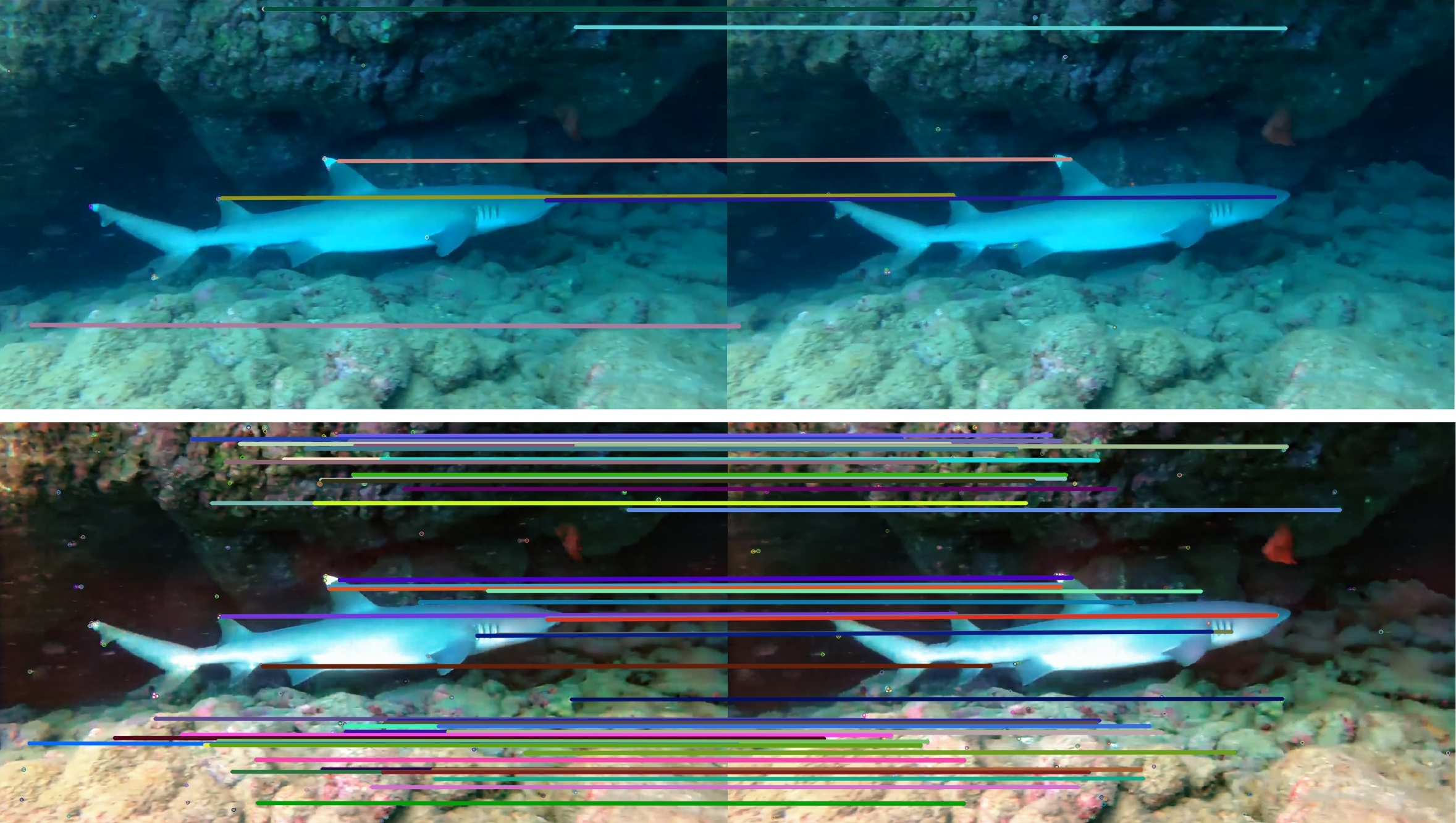}
    } \\ 
    \subfigure[Number of Inliers for sequential images]{%
        \includegraphics[trim = 105 340 115 350, clip, width=0.95\columnwidth]{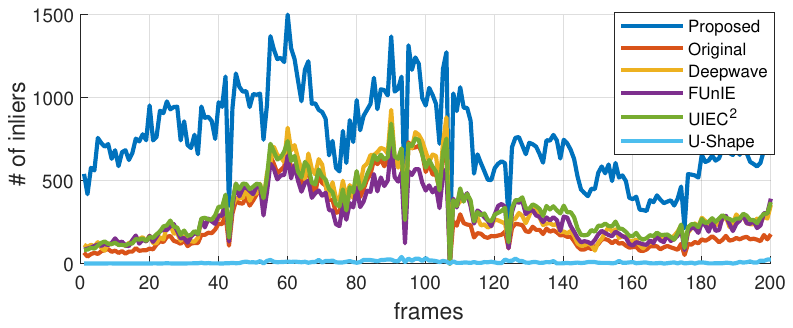}
    } \label{sub_fig_feat_match} 
    \vspace{-2mm}
    \caption{Feature matching results on a real underwater video.  (a) Qualitative sample of feature matching before (left) and after (right) enhancement with \texttt{Joint-ID}. (b) Number of feature-matching inliers for sequential images.}
    \label{fig:app_matching}
\vspace{-5mm}
\end{figure}

\subsection{Saliency Detection}

A multi-task operation is simultaneously performed in SESR \cite{islam2020simultaneous} for enhancement, super-resolution, and saliency detection. To evaluate saliency detection performance, we applied a saliency detection branch model, based on the SESR \cite{islam2020simultaneous}. In \figref{fig:app_saliency}, the result shows that the yellow dashed rectangles mark areas where important differences can be observed between the results of SESR and ours.

Also, our joint model detected more features in the area enclosed by the rectangles because it can further enhance the space along with depth estimation. This demonstrates the importance of understanding \ac{UIE} in underwater semantics and the advantages of simultaneously conducting depth estimation and image enhancement, as is done in our joint model.

\begin{figure}[h!] %
    \begin{center}
        \includegraphics[trim = 0 15 0 0, clip, width=1\columnwidth]{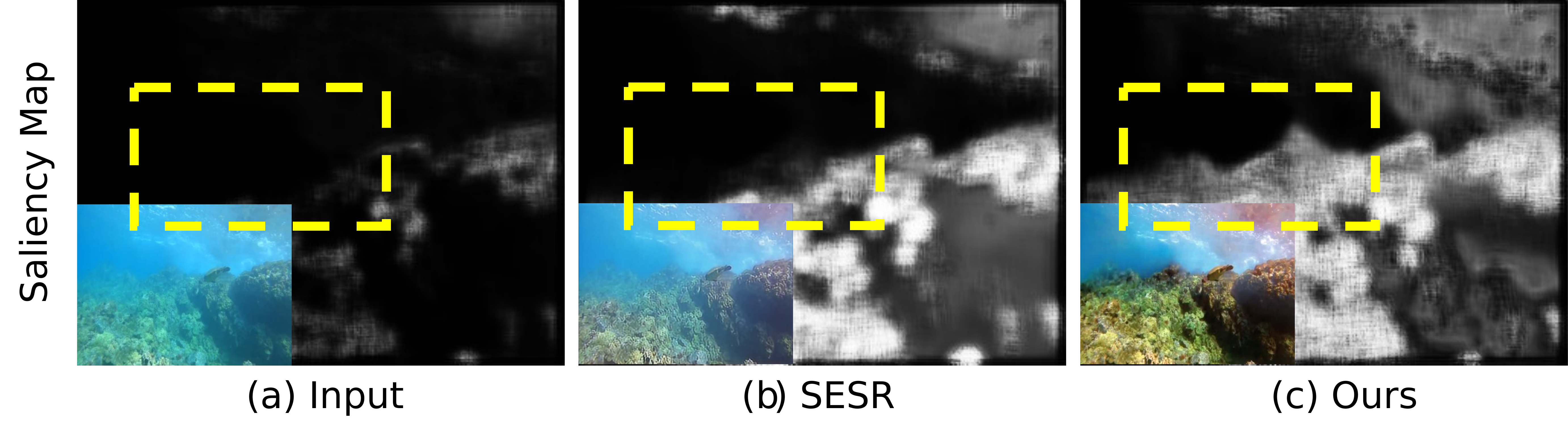}
        \vspace{-4mm}
        \caption{Qualitative evaluation of saliency detection between SESR \cite{islam2020simultaneous} and our model on a real underwater image sample.}
        \label{fig:app_saliency}
    \end{center}
    \vspace{-7mm}
\end{figure}

\section{Conclusion \& Future Work}

In this study, we presented a new learning-based solution for 3d underwater robot vision that addresses the need for simultaneous enhancement and depth estimation when interpreting underwater images. Most importantly, by designing joint blocks that share each task features, we demonstrated that a learning approach that targets the objectives of enhancement and depth estimation simultaneously is more powerful and effective than one that targets them separately, and presented a detailed description of the novel network architecture, called \texttt{Joint-ID}, based the approach and end-to-end training pipeline. Furthermore, we created a synthetic dataset of approximately 48000 labeled samples to facilitate training. Finally, we performed extensive qualitative and quantitative evaluations, which demonstrate that the proposed model: i) achieves \ac{SOTA} performance on \ac{UIE} and \ac{UMDE}; ii) exhibits significantly better generalization performance on standard datasets and real underwater images than other existing models; iii) provides high quality results on terrestrial images; and iv) effectively improves underwater feature matching and saliency prediction for underwater 3D robot vision. In the future, the subsequent study  will utilize the depth information for 3D robotics tasks (3D reconstruction, 3D object detection, etc.). Also, we aim to incorporate this mechanism into an underwater RGB-D SLAM system with an efficient depth uncertainty module and lightweight network.

\section*{Acknowledgment}
This work was supported by Inha University and the National Research Foundation of Korea(NRF) grant funded by the Korea government(MSIT) (No. RS-2023-00302589 and No.2022R1A4A3029480). Also, this work is supported by Institute of Information \& communications Technology Planning \& Evaluation (IITP) grant funded by the Korea government(MSIT) (No.2022-0-00448) and  Korea Institute of Marine Science and Technology Promotion (KIMST), funded by the Ministry of Oceans and Fisheries (20210562).

\small 
\bibliographystyle{packages/IEEEtranN} %
\bibliography{packages/string-short, packages/references}

\begin{IEEEbiography}[{\includegraphics[width=1in,height=1.25in,clip,keepaspectratio]{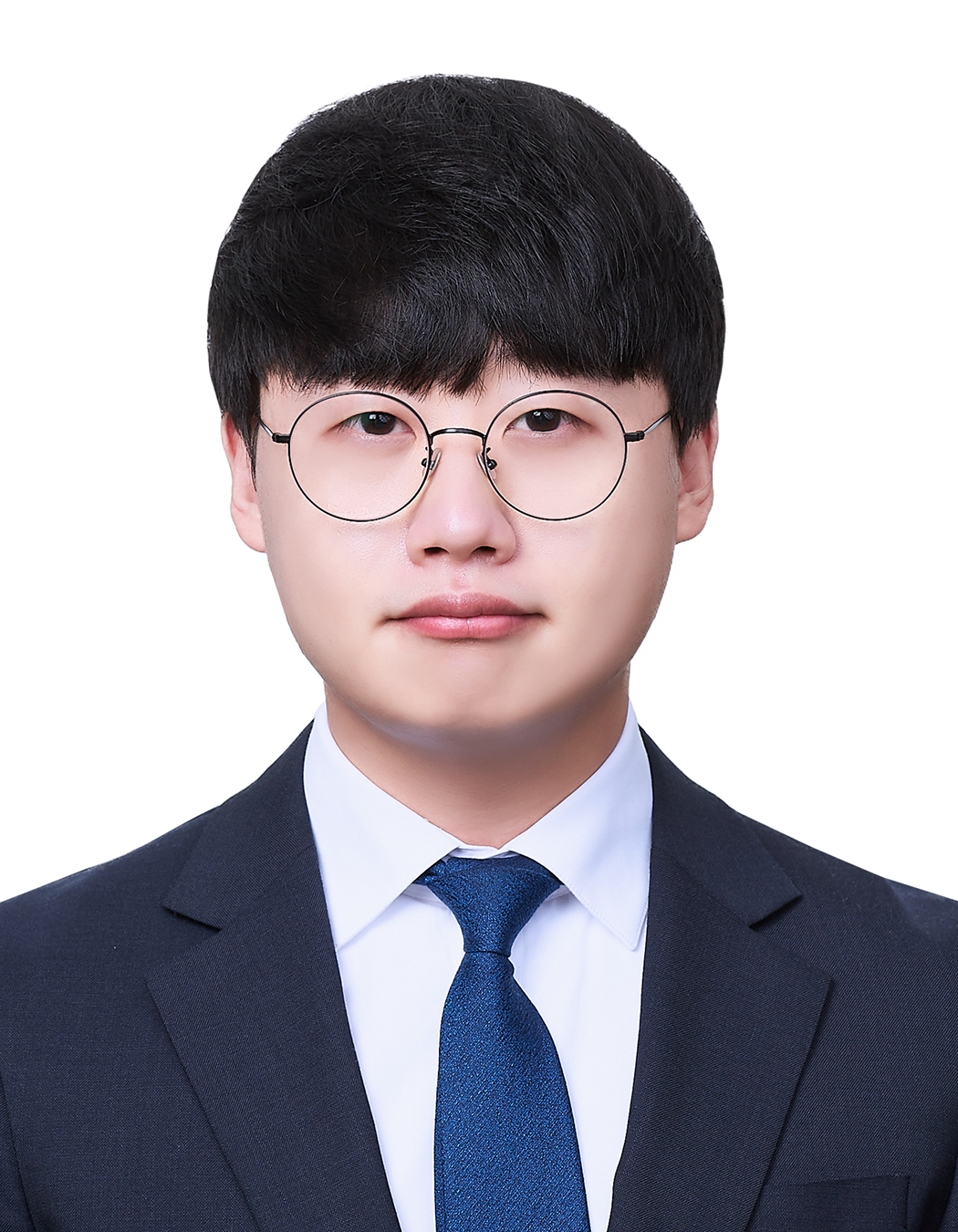}}]{Geonmo Yang} (S'22) received the B.S. degree in the electronic engineering from Tech University of Korea (TUKOREA), Siheung, South Korea, in 2022. Currently, he is M.S. student in the electrical and computer engineering from Inha University, Incheon, Korea. His research interests include visual simultaneous localization and mapping (SLAM), deep learning for computer vision and robotics for long-term autonomy.
\end{IEEEbiography}

\begin{IEEEbiography}[{\includegraphics[width=1in,height=1.25in,clip,keepaspectratio]{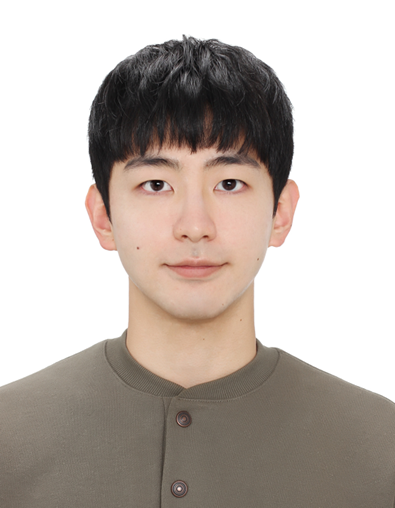}}]{GILHWAN KANG} (S'22) received the B.S. degree in the electrical engineering in the Inha University, Incheon, South Korea, in 2023. She is currently a M.S. student in the electrical and computer engineering from Inha University. His research interests include simultaneous localization and mapping (SLAM) and deep learning for 3D scene understanding.
\end{IEEEbiography}

\begin{IEEEbiography}[{\includegraphics[width=1in,height=1.25in,clip,keepaspectratio]{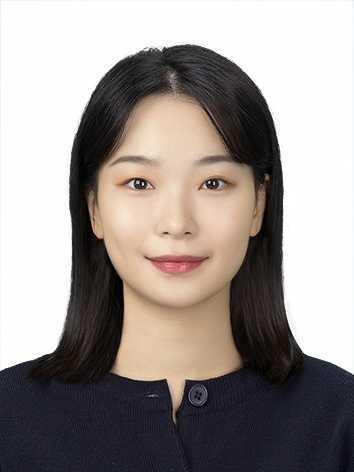}}]{JUHUI LEE} (S'22) received the B.S. degree in the electrical engineering in the Inha University, Incheon, South Korea, in 2022. She is currently a M.S. student in the electrical and computer engineering from Inha University. Her research interests include visual simultaneous localization and mapping (SLAM) and deep learning for computer vision.
\end{IEEEbiography}

\begin{IEEEbiography}[{\includegraphics[width=1in,height=1.25in,clip,keepaspectratio]{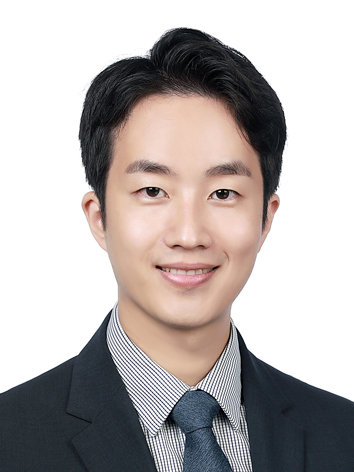}}]{Younggun Cho} (S'15--M'20) received the B.S. degree in electrical engineering from Inha University, Incheon, Korea, in 2013, and the M.S. degree in electrical engineering in 2015 and Ph.D. degree in civil and environmental engineering with dual degree in robotics program (2020) from the Korea Advanced Institute of Science \& Technology (KAIST), Daejeon, Korea. He was an Assistant Professor with the department of Robotics Engineering at Yeungnam University. He is currently an Assistant Professor with the department of Electrical Engineering at Inha University, Inchdeon, South Korea. His current research interests include robust sensing, long-term autonomy, and visual simultaneous localization and mapping.
\end{IEEEbiography}

\end{document}